%% file: paper.tex
\documentclass{Interspeech2024}
\usepackage{arydshln}

\interspeechcameraready

\title{Knowledge boosting during low-latency inference}

\name{Vidya Srinivas,$^1$ Malek Itani,$^1$ Tuochao Chen,$^1$ Sefik Emre Eskimez,$^2$ Takuya Yoshioka,$^3$ \\ Shyamnath Gollakota$^1$ }

\address{$^1$University of Washington, $^2$Microsoft, $^3$AssemblyAI}
\email{\{vysri,malek,tuochao\}@cs.washington.edu, sefik.eskimez@microsoft.com, takuya.yoshioka@ieee.org, gshyam@cs.washington.edu}


\begin{document}

\maketitle

\input{abstract-2}

\input{intro-3}

\input{related-3}

\input{methods-2}

\input{results-3}

\input{conclusion-2}

\bibliographystyle{IEEEtran}
\bibliography{mybib}

\end{document}

%% file: abstract-2.tex
\begin{abstract}
Models for low-latency, streaming applications could benefit from the knowledge capacity of larger models, but edge devices cannot run these models due to resource constraints. A possible solution is to transfer hints during inference from a large model running remotely to a small model running on-device. However, this incurs a communication delay that breaks real-time requirements and does not guarantee that both models will operate on the same data at the same time. We propose knowledge boosting, a novel  technique that allows a large model to operate on time-delayed input during inference, while still boosting small model performance. Using a  streaming neural network that processes 8~ms chunks, we evaluate different speech separation and enhancement tasks with communication delays of up to six chunks or 48~ms. Our results show larger gains where the performance gap between the small and large models is wide, demonstrating a promising method for large-small model collaboration for low-latency applications. 

\noindent {Code, dataset, and audio samples available at {\textcolor{blue}{{{\url{https://knowledgeboosting.cs.washington.edu/}}}}}}
\end{abstract}


%% file: intro-3.tex
\section{Introduction}

Advancements in deep learning, hardware, and algorithms have enabled models to run on diverse devices, from wearables to GPU clusters. While some small models can run on-device, large models require remote servers or the cloud. Resource-constrained applications can greatly benefit from the knowledge capacity of larger models, but cannot easily utilize these models during inference. We pose the following question: Can a remote large model boost the performance of an on-device small model during low-latency inference? An affirmative answer would benefit real-time applications across various domains such as robotics, self-driving vehicles, and audio and video processing.

\begin{figure}[t]
  \centering
  \includegraphics[width=0.8\linewidth]{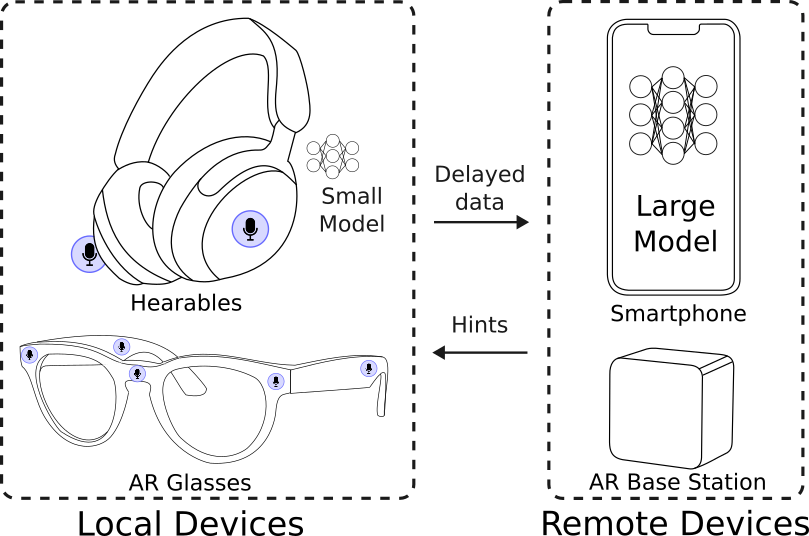}
  \caption{Example use cases for knowledge boosting. During inference, small models running locally can benefit from the knowledge capacity of large models running remotely. In these examples, the communication latency between the local and remote devices can exceed the  real-time processing requirements.}
  \vskip -0.15in
  \label{fig:applications}
\end{figure}

In this paper, we explore this question in the context of hearables and augmented audio applications, shown in Fig.~\ref{fig:applications}, targeting real-time speech manipulation tasks such as target speech extraction~\cite{tse,10.1145/3613904.3642057}, speech enhancement~\cite{dubey2023icassp,clearbuds}, and blind source separation~\cite{speechsep}. Such applications have low latency requirements that demand real-time, streaming processing of small chunks of audio ($\leq 10$ ms) and must operate on-device with limited computational resources. This requires models with minimal parameters and computational footprint~\cite{semantichear,waveformer,earbudsharing}. Given these constraints, we investigate if a large model running, for instance, on a nearby smartphone, can boost the inference-time performance of a small model running on a wearable.

A key challenge is that communication latency between two devices can exceed real-time processing requirements. Data wirelessly transmitted to a remote device introduces a delay between current inputs of the on-device small model and the remote large model. For example, according to the Bluetooth 5.0 standard~\cite{bt5}, the minimum delay is 15~ms roundtrip, but this can increase depending on wireless network congestion and interference. As a result, low-latency audio Bluetooth chips (e.g., Qualcomm aptX) guarantee only 40 ms latency~\cite{qualcomm}. Thus, the remote large model must operate on time-delayed input without access to the current audio chunks.

In this paper, we introduce \textit{knowledge boosting}, a novel technique in which delayed hints are provided by a large model to a small model during low-latency inference. Knowledge boosting enables a small model to accept hints after a time delay from a larger model, boosting its performance. Our key insight is that delayed large model information, when aligned with relevant history, can still enhance the current small model output. Further, through joint training, the large model can learn to provide useful hints that can improve real-time performance.

We evaluate our approach with very small models (around 40k parameters) that can fit on-device for wearables and large models (around 500k parameters) that can fit on-device on smartphones~\cite{semantichear}. We test the performance of our models on three binaural audio tasks, namely blind source separation (SS), speech enhancement (SE), and target speech extraction (TSE), using a streaming version of TF-GridNet~\cite{tfgridnet}. We also analyze our technique with ablation studies, through training configurations, compression ratios, and delays.   Our results demonstrate that,  at a delay of 48~ms (or six audio chunks), knowledge boosting improves the scale-invariant signal-to-distortion-ratio (SI-SDR) for SE, SS, and TSE by 0.23, 2.31, and 3.53 dB, respectively, over the corresponding vanilla models with a similar number of parameters as our on-device models. Our results demonstrate that the improvement from knowledge boosting depends on the performance gap between small and large models, with  large gains observed for TSE,  and smaller gains for SE.

%% file: related-3.tex

\begin{figure}[t!]
  \centering
  \includegraphics[scale=0.41]{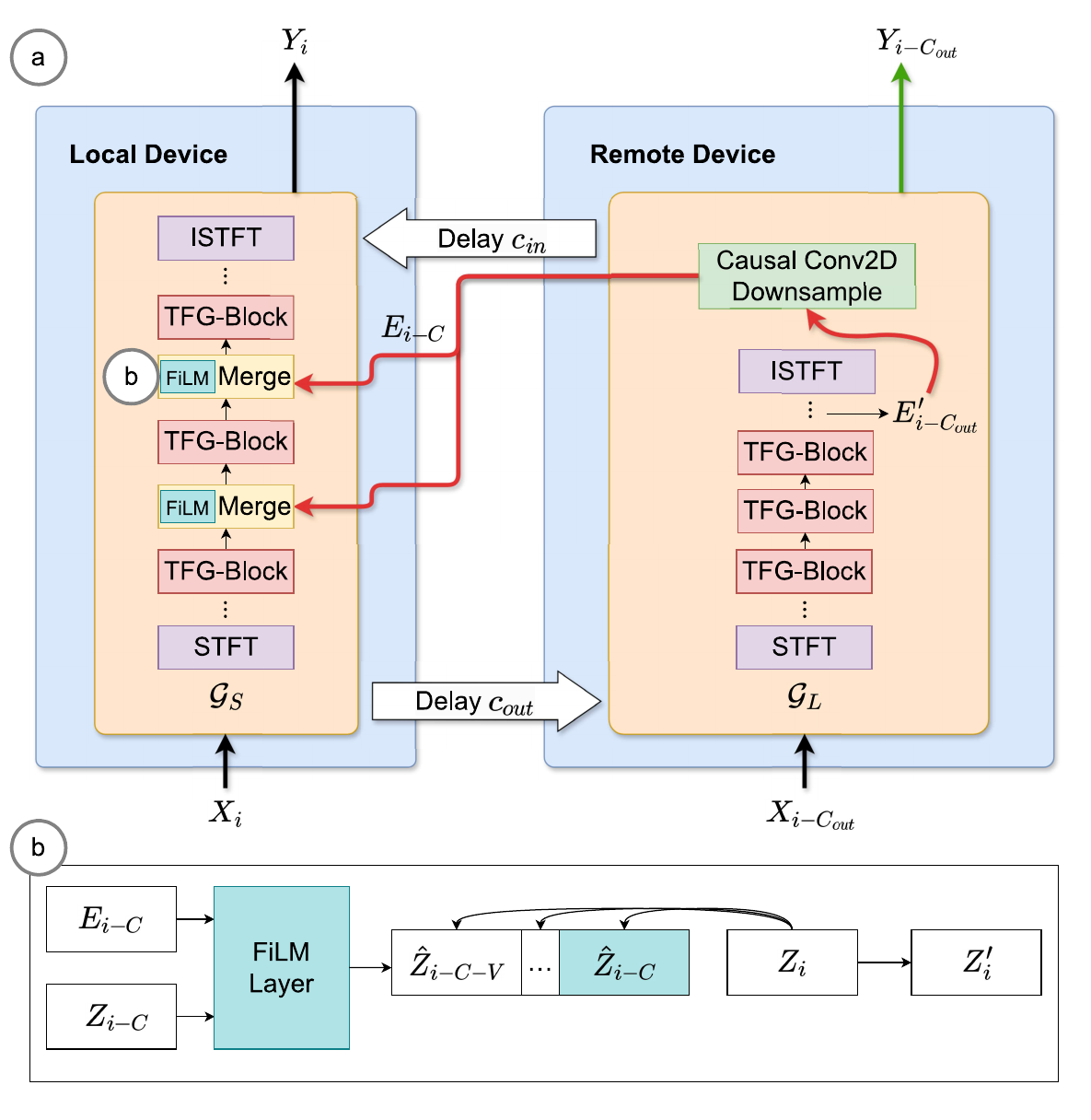}
  \vskip -0.15in
  \caption{Our system architecture. The green arrow is present only during large model pre-training.  The red arrows are present  during knowledge boosting.  The black arrows are present both during pre-training and knowledge boosting. }
  \label{fig:system_overview}
  \vskip -0.2in
\end{figure}

\section{Related Work}

\noindent{\bf Model partitioning:} This involves taking one large model and determining an optimal partitioning point such that one part of the model runs on a smaller device, and the other part runs on the cloud or a larger device. Prior works~\cite{part-li-aug2018, part-zeng-dec2021} have proposed model partitioning to adapt models to resource-constrained environments, executing only the necessary computation on-device and offloading the rest of the computation to the cloud. In this setting, both models have access to the same input at the same time. In contrast, our work aims to utilize a representation of the knowledge contained in a large model and use it to assist a small model after a delay. More importantly, because of communication and inference delay, the large model does not have access to the current small model input and has to work on previously received input samples.

\noindent{\bf Retrieval during inference and speculative decoding:}  In retrieval-augmented knowledge distillation~\cite{reaugkd-zhang2023}, a student and teacher model are trained jointly to minimize divergence between their probability distributions. Embeddings from the teacher model are frozen into a database. The student model uses its output to look up the related embedding from the teacher database to assist during inference. Another technique, speculative decoding, provides a prompt to a draft student model and then uses a larger teacher model for verification~\cite{decoding-leviathan2023, decoding-liu2023, decoding-yan2024}. The student model provides a set of proposals, or distributions for the target task. These proposals are then confirmed or denied by the large model. While the idea of using teacher knowledge to enhance student performance is similar to our work, these techniques do not take into account communication delay between the small and large model, the need for low latency inference, or model viability on small devices.

\noindent{\bf Knowledge distillation:} This technique, first proposed in~\cite{kd-hinton2015} transfers the knowledge of a large model to a small model during training for classification tasks. Variants of the original knowledge distillation proposal have been proposed for regression-based tasks~\cite{kd-takamoto2020}, and more specifically for speech and audio tasks~\cite{kd-chen2018, kd-hao2020,nathoo2023twostep}. In contrast to these works, which distill knowledge during training, knowledge boosting transfers representations between a small and large model during inference time for boosting the small model's performance. 

%% file: methods-2.tex
\section{Knowledge Boosting}
\label{section:methods}

\subsection{Problem formulation} 

Knowledge boosting utilizes two models --- a small and a large model. The small model receives chunks of binaural audio chunks $X_1, \dots, X_i$, where $X_{i} \in \mathbb{R}^{2 \times \tau f_s}$, $\tau$ is the chunk duration in seconds, and $f_s$ is the sampling rate. After receiving a chunk $X_i$, the local device sends it to the remote device, where it is received after a communication delay of $c_{out}$ seconds, or  $C_{out}$ chunks. The remote device processes this chunk with a neural network $\mathcal{G_{L}}$, computes an embedding $E_{i}$, and transmits it back to the local device. The embedding reaches the local device after a communication delay of $c_{in}$ seconds, or $C_{in}$ chunks. As a result, the embedding computed from the chunk $X_i$ arrives back at the local device after a delay, $c = c_{in} + c_{out}$. During this time, the local device would have received $C = \lfloor \frac{c_{in}+ c_{out}}{\tau}\rfloor$ additional chunks. Given our low-latency streaming requirements, the small model, through  $\mathcal{G_S}$, must produce an output chunk $Y_i$ while only using the information available to it by the time of input chunk, $X_i$, namely, the input chunks $X_{1}, \dots, X_{i}$ and the embeddings from the large model, $E_{1}, \dots, E_{i-C}$.

\subsection{System architecture} 
We design our network architecture using the  multi-channel causal TF-GridNet implementation~\cite{tfgridnet-ha}. Specifically, we use this network for both the small and the large models, $\mathcal{G_{S}}$ and $\mathcal{G_L}$.  Each network takes a time-domain binaural audio signal $x \in \mathbb{R}^{2 \times t}$ of length $t$ samples and uses a short-time Fourier transform (STFT) to convert it to a time-frequency representation $S \in \mathbb{C}^{2 \times F \times T}$, where $F$ is the number of frequency bins and $T$ is the number of time frames. Then, a $D$-dimensional latent representation $Z \in \mathbb{R}^{D \times F \times T}$ is generated and processed with a sequence of TF-GridNet blocks, where the output of the $j$-th TF-GridNet block is $Z^{j}$. The output of the last TF-GridNet block is then mapped to  $K$ time-frequency domain channels, $\hat{S} \in \mathbb{C}^{K \times F \times T}$, and converted back to the time domain using an inverse STFT. Further details can be found in \cite{tfgridnet}.

The large model generates embeddings that are received by the small model during training and inference. The embeddings are generated from the intermediate representation, $\hat{S}$, right before the inverse STFT in the TF-GridNet model. Specifically, we concatenate the real and imaginary components along the channel dimension, $K$, to generate the embedding, $E' \in \mathbb{R}^{2K \times F \times T}$. This embedding is passed through a compression module, which takes an input $E'$ and outputs $E\in \mathbb{R}^{2K/P \times F \times T}$ where $P$ is the compression ratio. The compression module is implemented as a single casual convolution layer with a kernel size of 3. The outputs of the compression module are then passed to the small model.

For a given chunk $X_i$, the small model computes an input representation $Z^0_i$  which it passes through a sequence of  TF-GridNet blocks. The small model incorporates the compressed embeddings coming from the large model via the merge modules, shown in Fig.~\ref{fig:system_overview}b. These modules are located in between consecutive TF-GridNet blocks. The $j$-th merge module takes two inputs--- the latent representation, $Z^j$, from the output of the $j$-th TF-GridNet block and a time-delayed embedding, $E_{i-C}$ from the large model. We use a context length $V$ in our merge modules. Specifically, given the large model embedding, $E_{i-C}$, and the latent representation, $Z^j_{i-C}$, we compute the contextual representation, $\hat{Z}^j_{i-C}$,  using a FiLM layer~\cite{film}. We then compute the merged output, ${Z}^{'j}_i$, using multi-head cross attention between $[\hat{Z}^j_{i-C-V},\cdots, \hat{Z}^j_{i-C}]$ and $Z^j_i$. ${Z}^{'j}_i$ is then provided as input to the $(j+1)$-th TF-GridNet block. During inference, to minimize computational complexity, we cache the last $C+V$ contextual representations computed previously. 

\subsection{Training procedure} 
We first pre-train the large model on the target task, which yields a reasonable initial set of weights. To train for knowledge boosting, we first process an input audio sequence with the large model to obtain the embeddings $E_1, \cdots, E_N$, where $N$ is the total length of the audio sequence. Then, we simulate the communication latency by time-shifting the embedding sequence to the right by $C$, feeding in zeros to the first time frames where no embedding is available, and passing this sequence to the small model along with the original chunks $X_1, \dots, X_N$. Both models are jointly trained, and the small model's output is used to compute the loss function for training. During backpropagation, we update the parameters of both the large and small models.

%% file: results-3.tex
\section{Experiments and Results}

\noindent{\bf Datasets.} To generate binaural audio mixtures, we first sampled rooms from one of four binaural room impulse response datasets -- CIPIC~\cite{cipic}, RRBRIR~\cite{rrbrir}, ASH-Listening-Set~\cite{ash-listening}, and CATTRIR~\cite{CATT_RIR} -- with probability 0.35, 0.05, 0.45 and 0.15. We then sampled  speech utterances from the LibriSpeech~\cite{LibriSpeech} dataset and convolved each of them with a binaural room impulse response from the sampled room. We summed up these binaural speech signals to obtain a binaural speech mixture. For the SS and TSE tasks, the resulting mixture was created from two binaural speech utterances. For SE, we only used a single speech utterance. Instead of using the room impulse response on a single-channel noise signal, which would only simulate a noise source in a single direction, we used a binaural noise signal recorded in the WHAM!~\cite{wham} dataset as our ambient binaural noise. We scaled the noise so that the resultant average signal-to-noise ratio across both microphone channels is uniformly distributed between $[-6,6]~$dB. For each task, we generated 100,000 mixtures for training, 5,000 for testing, and 5,000 for validation. Speech utterances were sampled from LibriSpeech's \verb|train-clean-360|, \verb|test-clean| and \verb|dev-clean|, respectively. Noise samples were sampled from WHAM!'s \verb|tr|, \verb|tt|, and \verb|cv|,  respectively. All mixtures were 5~s long and the  sampling rate was 16~kHz. There was no overlap in the identity of the speakers and the noise sources between the train, test, and validation splits.\vskip 0.02in


\begin{table}[t!]
  \caption{Main results with  delay of 48~ms ($C=6$). The prefixes ``Mx", and ``KB" refer to the input mixture and knowledge boosting, respectively. The prefixes "S", "M", and "L" refer to the small, medium, and large model baselines, respectively.``Param." specifies the number of model parameters and ``MACs" the number of multiply-accumulate operations. The MACs reported for KB configurations are for the {small} model only, as the small model is the only part that must run on a local device. For TSE, the speaker embedding parameters are excluded since they only need to be computed once and cached; this computation can occur on the remote device.}
  \label{tab:main-res}
  \vskip -0.1in
  \centering
\setlength{\tabcolsep}{5pt}
 
  \begin{tabular}{ c c c c c c }
    \toprule
    \multicolumn{1}{c}{\textbf{Name}} & 
    \multicolumn{1}{c}{\textbf{SI-SDR}} &
    \multicolumn{1}{c}{\textbf{PESQ}} &
    \multicolumn{1}{c}{\textbf{STOI}} & \multicolumn{1}{c}{\textbf{Param.}} &
    \multicolumn{1}{c}{\textbf{MACs}}\\
    
    & 
    \textbf{(dB)}&
    &
    &
    \textbf{(K)}&
    \textbf{(M)}\\
    \midrule
    Mx-SS & 0.00 & 1.28 & 0.70 & - & -\\
    \hdashline[1pt/2pt]\hdashline[0pt/1pt] 
    M-SS & 9.72 & 2.05 & 0.85 & 37.38 & 3.70\\ 
    KB-SS & \textbf{12.03} & \textbf{2.23} & \textbf{0.87} & 36.54 & 2.68\\
    \hdashline[1pt/2pt]\hdashline[0pt/1pt] 
    S-SS & 8.65 & 1.93 & 0.84 & 23.96 & 2.40\\ 
    L-SS & 13.92 & 2.59 & 0.89 & 518.77 & 37.98\\
    \midrule
    Mx-SE & 0.01 & 1.12 & 0.67 & - & -\\
    \hdashline[1pt/2pt]\hdashline[0pt/1pt] 
    M-SE & 9.34 & 1.75 & 0.82 & 36.44 & 3.61\\
    KB-SE & \textbf{9.57} & \textbf{1.77} & \textbf{0.83} & 35.70 & 2.60\\
    \hdashline[1pt/2pt]\hdashline[0pt/1pt] 
    S-SE & 9.01 & 1.71 & 0.82 & 23.38 & 2.35\\ 
    L-SE & 11.28 & 2.12 & 0.87 & 516.46 & 37.76\\
    \midrule
    Mx-TSE & 1.05 &  1.36 & 0.72 & - & -\\
    \hdashline[1pt/2pt]\hdashline[0pt/1pt]
    M-TSE & 4.00 & 1.45 & 0.75 & 36.44 & 6.19\\
    KB-TSE & \textbf{7.53} & \textbf{1.92} & \textbf{0.82} & 35.70 & 4.19\\ 
    \hdashline[1pt/2pt]\hdashline[0pt/1pt] 
    S-TSE & 3.95 & 1.42 & 0.75 & 23.38 & 3.94\\ 
    L-TSE & 8.52 & 2.21 & 0.84 & 516.46 & 44.12\\
    \bottomrule
  \end{tabular}
    \vskip -0.2in
\end{table}


\noindent{\bf Evaluation Setup.} We  compared  the performance of a small model augmented with delayed hints from a large model during inference time with a vanilla model of a similar parameter size. Following the naming conventions in~\cite{tfgridnet}, $D$ is the embedding dimension for each TF unit, $B$ is the number of TF-GridNet blocks, $I$ is the kernel size for unfold and Deconv1D, $J$ is the stride size for unfold and Deconv1D, $H$ is the number of hidden units in the LSTM layers and $L$ is the number of heads in self-attention. In all our experiments, we set $I=J=1$. We computed the time-frequency representations using uncentered 12~ms STFT windows with a hop size of $8$~ms. We considered three baseline models per task. We trained a baseline small model with $D=16, L=4, B=3, H=16$, and no attention. We also trained a baseline medium model with $D=26, L=4, B=3, H=18$, and no attention. Finally, we trained a baseline large model with $D=64, L=8, B=3, H =64$ with attention. We limit self-attention to the last 50 chunks. After establishing these baselines, we  trained knowledge boosting with a joint model training method. Specifically, we jointly trained a large model, with the same hyperparameters as our baseline large model, to boost the performance of a small model with the same hyperparameters as our baseline small model. In all experiments, we initialized the large model with the baseline large model weights, while the small model was not pre-trained. We use a context length of $V=49$ for cross-attention modules.\vskip 0.02in

\noindent{\bf Loss functions and training hyperparameters.} For the TSE and SE tasks, we optimized the network parameters to maximize the average scale-invariant signal-to-distortion ratio (SI-SDR)~\cite{sisdr} across the two microphone channels. For SS, we output two speaker channels per microphone, and used  permutation invariant training to maximize the average SI-SDR across speakers and microphones. In all three tasks, since we were not particularly concerned about binaural cues, we treated the left and right channels independently when computing the optimal scale factor for SI-SDR. We used this same scale factor to rescale when computing PESQ and STOI. For all experiments, in each epoch, we iterated over the entire training and validation sets, and halved the learning rate if the average scale-invariant signal-to-noise ratio (SI-SNR) over the validation set does not decrease after four iterations. For all training runs, we used a batch size of 8 and  gradient clipping with the norm set to 1. Our baseline models were trained for 100 epochs with an initial learning rate of 2e-3. Our main experiments were trained until convergence of the loss function with an initial learning rate of 1e-3. All of our ablation experiments were trained for 20 epochs. We used the Adam optimizer. In our evaluations, we used the best-performing weights on the validation set. \vskip 0.02in

\noindent{\bf Main results.} We tested the viability for knowledge boosting at a delay of 48 ms, or $C = 6$, and trained with a joint configuration, as specified above, for SS, SE, and TSE. We compared the results of a small model trained with knowledge boosting and a vanilla medium model of similar parameter size in Table~\ref{tab:main-res}. Overall, we observed that knowledge boosting tends to improve the performance of a small model over a vanilla model of a similar parameter size without knowledge boosting. We do so with a notable reduction in MACs. At delay $C=6$, we achieved a relative improvement of 0.23, 2.31, and 3.53 dB for SE, SS, and TSE, respectively, over the vanilla medium models for each task. A paired t-test was conducted for each task, showing a significant difference with $p<0.05$.  As compared to the vanilla medium models for SE, SS, and TSE, respectively, we achieved a 1.01 M, 1.02 M, and 2.00 M reduction in multiply-accumulates (MACs) using knowledge boosting. This performance improvement despite MAC reduction was due to the fact that we cut down the length of computationally expensive units, such as LSTMs, reduced the dimensionality of embeddings used in TF-GridNet, and replaced these ``missing" parameters with merge modules and delayed, but valuable hints from the large model.

\begin{table}[t]
  \caption{Knowledge boosting ablation experiments on SS for different delay configurations (C) with no compression. The prefixes ``KB", and ``FKB" refer to knowledge boosting and knowledge boosting with a frozen large model, respectively.}
  \vskip -0.1in
  \label{tab:ablation-ss-fm-d}
  \centering
  \begin{tabular}{ c c c c c }
    \toprule
    \multicolumn{1}{c}{\textbf{Name}} & \multicolumn{1}{c}{\textbf{C}}  &
    \multicolumn{1}{c}{\textbf{SI-SDR (dB)}} & \multicolumn{1}{c}{\textbf{PESQ}} &
    \multicolumn{1}{c}{\textbf{STOI}}\\

    \midrule

    FKB-SS & 0 & \textbf{13.50} & \textbf{2.47} & \textbf{0.89} \\
    KB-SS & 0 & 13.11 & 2.45 & \textbf{0.89} \\
    \hdashline[1pt/2pt]\hdashline[0pt/1pt] 
    FKB-SS & 1 & 10.73 & 2.04 &  0.87 \\
    KB-SS & 1 & \textbf{11.88} & \textbf{2.23} & \textbf{0.88} \\
    \hdashline[1pt/2pt]\hdashline[0pt/1pt] 
    FKB-SS & 3 & 9.77 & 1.89 & 0.84 \\
    KB-SS & 3 & \textbf{11.21} & \textbf{2.07} & \textbf{0.86} \\    
    \hdashline[1pt/2pt]\hdashline[0pt/1pt] 
    FKB-SS & 6 & 9.27 & 1.93 & 0.84 \\   
    KB-SS & 6 & \textbf{10.73} & \textbf{2.04} & \textbf{0.86} \\
     
    \bottomrule
  \end{tabular}
      \vskip -0.1in
\end{table}

\begin{table}[t]
  \caption{Knowledge boosting ablation experiments on SS for different delay values with different compression factors ($P$).}
  \vskip -0.1in
  \label{tab:ablation-ss-cmp}
  \centering
  \begin{tabular}{ c c c c c c c}
    \toprule
    \multicolumn{1}{c}{\textbf{Name}} & \multicolumn{1}{c}{\textbf{C}}  & \multicolumn{1}{c}{\textbf{P}} &
    \multicolumn{1}{c}{\textbf{SI-SDR}} & \multicolumn{1}{c}{\textbf{PESQ}} &
    \multicolumn{1}{c}{\textbf{STOI}} &
    \multicolumn{1}{c}{\textbf{MACs}}\\
    & 
    &
    &
    \textbf{(dB)}&
    &
    &
    \textbf{(M)}\\
    \midrule

    KB-SS & 0 & 2 & 13.17 & 2.46 & 0.89 & 2.655\\
    KB-SS & 0 & 4 & 12.75 & 2.43 & 0.89 & 2.643 \\
    \hdashline[1pt/2pt]\hdashline[0pt/1pt] 
    KB-SS & 1 & 2 & 11.98 & 2.25 & 0.88 & 2.655\\
    KB-SS & 1 & 4 & 11.13 & 2.08 & 0.87 & 2.643\\
    \hdashline[1pt/2pt]\hdashline[0pt/1pt] 
    KB-SS & 6 & 2 & 10.59 & 2.00 & 0.86 & 2.655\\
    KB-SS & 6 & 4 & 10.28 & 1.99 & 0.86 &  2.643\\
    \bottomrule
  \end{tabular}
    \vskip -0.15in
\end{table}

\begin{table}[t]
  \caption{Knowledge boosting ablation experiments on TSE and SE for different delay configurations (C) with no compression.}
  \vskip -0.1in
  \label{tab:ablation-tse-se}
  \centering
  \begin{tabular}{ c c c c c }
    \toprule
    \multicolumn{1}{c}{\textbf{Name}} & \multicolumn{1}{c}{\textbf{C}}  &
    \multicolumn{1}{c}{\textbf{SI-SDR (dB)}} & \multicolumn{1}{c}{\textbf{PESQ}} &
    \multicolumn{1}{c}{\textbf{STOI}}\\
    \midrule
    KB-SE & 0 & 11.24 &  0.87 & 2.11 \\
    KB-SE & 1 & 10.64  & 0.85 & 1.95 \\
    KB-SE & 3 & 9.88  & 0.84 & 1.82 \\
    KB-SE & 6 & 9.57  & 0.83 & 1.77 \\
    \midrule
    KB-TSE & 0 & 9.34 &  0.84 & 2.31 \\
    KB-TSE & 1 & 8.20  & 0.83 & 2.05 \\
    KB-TSE & 3 & 7.65  & 0.82 & 1.86 \\
    KB-TSE & 6 & 7.70  & 0.82 & 1.89 \\
    \bottomrule
  \end{tabular}
  \vskip -0.2in
\end{table}

\noindent{\bf Ablation studies.} We performed  ablation studies to evaluate the effects of large model weight freezing, compression, and different delays on the SS tasks, and evaluate the performance at different delays on the SE and TSE tasks. Specifically, for the SS task, we trained the large model and small model jointly, with a compression ratio $P=1$  in Table~\ref{tab:ablation-ss-fm-d}, and swept  delays $C = 0, 1, 3, 6$ chunks, corresponding to $0, 8, 24,$ and $48$ ms, respectively. In Table~\ref{tab:ablation-ss-fm-d}, we also investigated the effects of freezing the large model during training on overall performance, using $P=1$ and sweeping $C = 0, 1, 3, 6$ chunks. In Table~\ref{tab:ablation-ss-cmp}, we tested compression ratios $P= 1, 2, 4$ for $C = 0, 1, 6$ chunks. Finally, we also swept $C = 0, 1, 3, 6$ with $P=1$ across SE and TSE in Table~\ref{tab:ablation-tse-se}. 

Our results show performance degradation with larger $C$ values. We found that freezing the large model during training led to lower performance than a joinly-trained large model. We also found that  compression of the information sent from the large model to the small model slightly drops performance across $C$ values. For $C=6$, this drop was  only 0.14~dB (p $>$ 0.3) and 0.45~dB (p $>$ 0.001) at compression factors, $P=2$ and $P=4$, respectively, compared to no compression. 

%% file: conclusion-2.tex
\section{Discussion and Limitations}
Our work has a few limitations that present opportunities for future research.  The device running the small model has to continuously stream audio to the remote device. However, it has been shown that wearable devices can do this over Bluetooth Low Energy (BLE) and still run continuously for 40 hours with only a coin cell battery~\cite{clearbuds}. The compression module impacts throughput requirements for transmitting embeddings from the remote to the small device. For the TSE task, uncompressed embeddings require a data rate of 1.55~Mbps,  while a 
compression factor of 2 reduces it to 776~kbps. Notably, BLE  supports a maximum rate of 2~Mbps~\cite{bt5};  and, recent work demonstrated concurrent  streaming of compressed audio from seven  microphones over BLE~\cite{acousticswarm}. Future work could explore more specialized compression techniques like neural vocoders~\cite{vocoder_codec} to further reduce throughput requirements.

We  trained different pairs of small and large models for each  communication delay. While it is possible to train a single pair of small and large models to accept variably delayed input, we defer this to future work along with exploring alternate merge methods.   Finally,  we explored  delays of up to 48~ms in the context of Bluetooth streaming from a wearable to a smartphone or a home base station. One could have larger  models running on the cloud, but transmission delays would be higher. However, such models could have a much larger capacity, potentially leading to applications for more complicated tasks; we leave this exploration to future research.

\section{Conclusion}

We proposed knowledge boosting, a novel technique for improving the performance of small models at inference time through  delayed hints from a large model. Our results show that knowledge boosting is a promising approach, worthy of further exploration,  for large-small model collaboration   during  low-latency streaming applications.

\section{Acknowledgments}
The UW researchers are funded by the Moore
Inventor Fellow award \#10617, UW CoMotion fund,  and the NSF.